\documentclass[10pt,twocolumn,letterpaper]{article}

\usepackage{cvpr}              

\usepackage{graphicx}
\usepackage{amsmath}
\usepackage{amssymb}
\usepackage{booktabs}

\usepackage[pagebackref,breaklinks,colorlinks]{hyperref}

\usepackage[capitalize]{cleveref}
\crefname{section}{Sec.}{Secs.}
\Crefname{section}{Section}{Sections}
\Crefname{table}{Table}{Tables}
\crefname{table}{Tab.}{Tabs.}

\def\cvprPaperID{*****} 
\def\confName{CVPR}
\def\confYear{2022}

\begin{document}

\title{A Strategy Optimized Pix2pix Approach for SAR-to-Optical Image Translation Task}

\author{Fujian Cheng\\
STAR.VISION\\
{\tt\small cheng.fujian@star.vision}
\and
Yashu Kang\\
STAR.VISION\\
{\tt\small kang.yashu@star.vision}
\and
Chunlei Chen\\
STAR.VISION\\
{\tt\small chen.chunlei@star.vision}
\and
Kezhao Jiang\\
Columbia University\\
{\tt\small kj2589@columbia.edu}
}
\maketitle

\begin{abstract}
   This technical report summarizes the analysis and approach on the image-to-image translation task in the Multimodal Learning for Earth and Environment Challenge (MultiEarth $2022$). In terms of strategy optimization, cloud classification is utilized to filter optical images with dense cloud coverage to aid the supervised learning alike approach. The commonly used pix2pix framework with a few optimizations is applied to build the model. A weighted combination of mean squared error and mean absolute error is incorporated in the loss function. As for evaluation, peak to signal ratio and structural similarity were both considered in our preliminary analysis. Lastly, our method achieved the second place with a final error score of $0.0412$. The results indicate great potential towards SAR-to-optical translation in remote sensing tasks, specifically for the support of long-term environmental monitoring and protection.
\end{abstract}

\section{Introduction}
\label{sec:intro}

Satellite remote sensing data has been widely used in various domains including urban management, agriculture, climate change, environmental monitoring\cite{chuvieco2010advances, anderson2017earth}, 
\etc. The recent advances in generation, enhancement and interpretation of RS imagery have helped gaining attentions of the machine learning community. On the other hand, the mass amount of available data collected from various sensors also poses challenges. For instance, Synthetic aperture radar (SAR) data is obtained through actively illuminating the ground with radio waves rather than utilizing reflectance values as with optical images, which is insensitive to lighting and weather conditions \cite{fuentes2019sar}.

For instance, in the Multimodal Learning for Earth and Environment Challenge (MultiEarth $2022$), a key component of this event is monitoring the Amazon rainforest regardless of weather and lighting conditions by utilizing a time series of multispectral and SAR images \cite{cha2022multiearth}. The data set covers Sentinel-1 synthetic aperture radar data, Sentinel-2 multispectral data, Landsat-5/8 multispectral data, etc. Such a multimodal data set offers new opportunities for a variety of applications as well as posing new challenges.

The goal of image-to-image translation track is to predict a set of possible cloud-free corresponding optical images given an input SAR image. Since rainy weather occurs to the Amazon rainforest for most of an entire year, continuous monitoring by optical imagery is extremely difficult. SAR has the advantage of invariance to weather and lighting conditions. Thus, the application of SAR-to-optical image translation are important topics when it comes to SAR interpretation. 

The image-to-image translation task is essentially transforming one kind of representation of a scenario into another. In recent years, some studies have utilized deep learning models on this task. Existing studies include pix2pix based on conditional generative adversarial networks (cGAN) \cite{isola2017image, bermudez2019synthesis}, cycle-consistent GAN (cycleGAN) \cite{turnes2020atrous,zhu2017unpaired}, \etc. Despite the downside of high computation cost, cGAN can reduce the speckle effect and produce relatively smooth textures. As for cycleGAN, it is more likely to generate hallucinations and d4eflected spatial accuracy compared with cGANs.



\subsection{Methodology}
\subsection{Data preparation}

\begin{figure}
  \centering
   \includegraphics[width=1\linewidth]{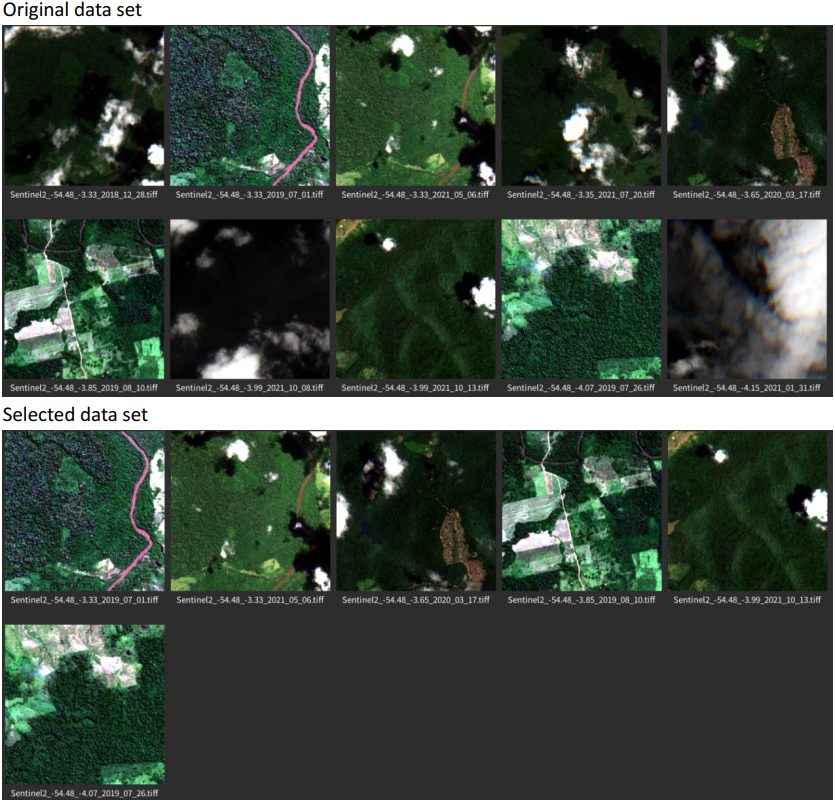}

   \caption{Data selection using cloud classification method}
   \label{fig:one}
\end{figure}

To fill the gaps in optical images and obtain interpretation, the goal is that the resultant optical images should also be free of dense cloud coverage. In our preliminary analysis, we used a simple VGG classification model to select data without presence of dense cloud. For instance, \cref{fig:one} demonstrates the outcome of the data selection results. In addition, in our $data\_loader$ module, pixel values are preprocessed by normalizing to $[-1, 1]$. 

\subsection{Model architecture}

\begin{figure}
  \centering
   \includegraphics[width=1\linewidth]{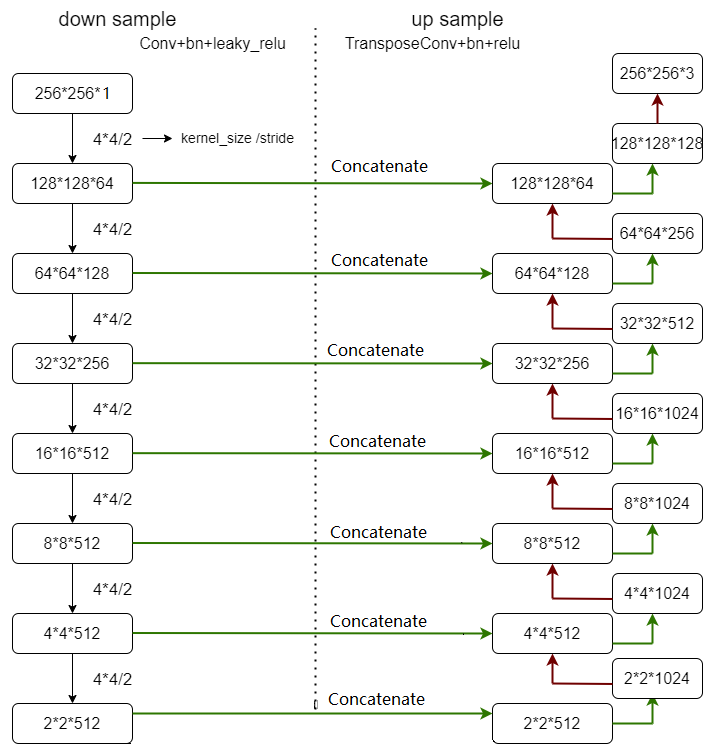}

   \caption{The U-net generator}
   \label{fig:two}
\end{figure}

\begin{figure}
  \centering
   \includegraphics[width=1\linewidth]{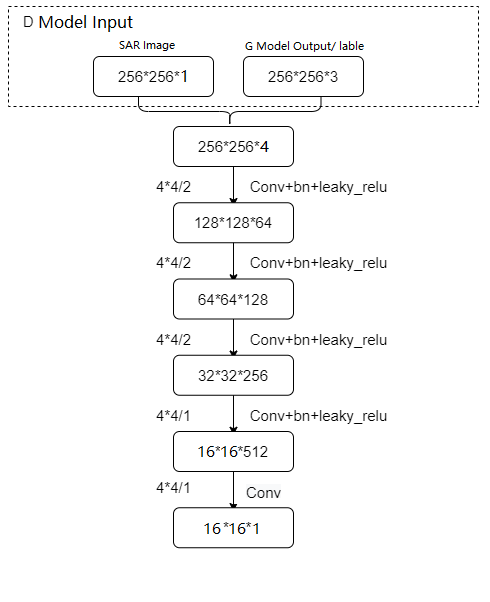}

   \caption{The CNN discriminator}
   \label{fig:three}
\end{figure}

Based on encoder-decoder framework, we utilize a classical pix2pix architecture where the generator follows a U-net design and Convolutional Neural Networks (CNN) discriminator, as shown in \cref{fig:two} and \cref{fig:three}. Using paired images as input, this modeling framework resembles a supervised learning task. Note that the data preparation described in the previous section helps guarantee that the objective image is not impacted by dense cloud.

The most commonly used pix2pix model incorporates two loss functions for training, conditional adversarial loss $L_{cGAN}$ and $L_{L1}$ which uses $L1$ norm to estimate difference with the target. This loss function considers both speckling effect and artifacts of the generated images. 

We also restricted the discriminator by performing backpropagation calculation every other training epoch to avoid an overpowered discriminator and slow convergence for the generator. As for the loss function design, a weighted combination of mean squared error (MSE) and mean absolute error (MAE) is used in this model.

\subsection{Evaluation metric}
Performance is evaluated based on error score\cite{cha2022multiearth}:
\begin{equation}
  \sum_{j}\min_{i}\|f(x)_{i}-y_{j}\|
  \label{eq:one}
\end{equation}
where $f(x)_{i}$ is a set of generated output images and $y_{i}$ is the corresponding optical image ground truth. That is, for each SAR image with two channels, there will be three possible outcomes.

\section{Analysis Results}
\label{sec:analysis}

\begin{figure}[h]
  \centering
   \includegraphics[width=1\linewidth]{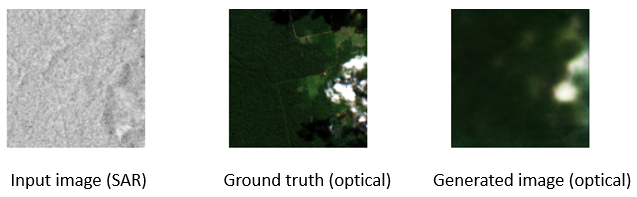}

   \caption{Experiments with uint16 image input}
   \label{fig:four}
\end{figure}

\begin{figure}[h]
  \centering
   \includegraphics[width=1\linewidth]{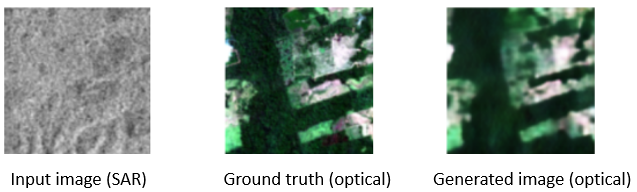}

   \caption{Experiments with uint8 image input}
   \label{fig:five}
\end{figure}

In our preliminary experiment, we trained the model with uint16 and uint8 images, respectively. The results are shown in \cref{fig:four} and \cref{fig:five}. It revealed that textual information and spatial information were better preserved. In addition, Peak signal to noise ratio (PSNR) and structural similarity (SSIM) were both calculated between the generated image and ground truth. 

PSNR, peak signal to noise ratio is a commonly used metric to define the similarity between two images. It is calculated using the MSE of the pixels and the maximum possible pixel value (MAX) as follows:

\begin{equation}
  PSNR=10\cdot\log\left({\frac{{MAX}^2}{MSE}}\right)
  \label{eq:two}
\end{equation}

SSIM, the structural similarity index (SSIM) is developed in order to take luminance, contrast and structure of both images into account. It is calculated on various windows of an image. The measure between two windows $x$ and $y$ of common size $N$ is:

\begin{equation}
  SSIM(x,y)=\frac{(2\mu_{x}\mu_{y} + c_{1})(2\sigma_{xy}+c_{2})}{(\mu_{x}^{2}+\mu_{y}^{2}+c_{1})(\sigma_{x}^{2}+\sigma_{y}^{2}+c_{2})}
  \label{eq:three}
\end{equation}

The uint16 experiment achieved average PSNR 20.734 and SSIM 0.735, while the uint8 experiment achieved average PSNR 21.551 and SSIM 0.763.

\begin{figure}[h]
  \centering
   \includegraphics[width=1\linewidth]{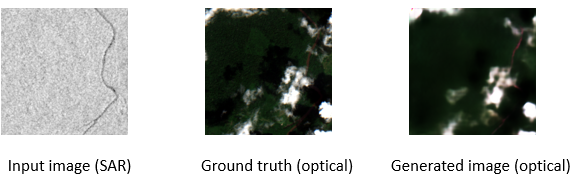}

   \caption{Final submission}
   \label{fig:six}
\end{figure}

Lastly, \cref{fig:six} demonstrates a sample from our final submission. Besides, the final submission achieved PSNR value $22.527$ and SSIM $0.732$. Besides, we found that implementing cumulative count cut processing was more efficient in producing more accurate output images and enhancing model convergence than optimization for uint16 input images. Due to time constraints, our final model achieved the second place with an error score of $0.0412$.








\section{Conclusions}

This technical report summarizes our efforts on SAR-to-optical image translation task in the MultiEarth $2022$ challenge. To generate accurate optical images from SAR input, we utilized cloud classification to filter cloud-dense optical images, so that the task is approached supervised-learning-like and the style of the generated images can be ensured. The deep learning model followed a general pix2pix framework and obtained a final error score of $0.0412$, achieving the second place. In terms of strategy optimization, the data preprocessing proved to be beneficial to help preserve relevant features in the original data set.

Overall, despite the availability of both SAR and optical data, the image translation task still posed a big challenge for us. However, the outcome indicates great potential towards SAR-to-optical translation in remote sensing tasks, specifically for the support of long-term environmental monitoring where dense cloud coverage is often present. Due to time constraints, we did not fully explore optimizations of other alternatives such as cGAN, cycleGAN or pix2pixHD, etc. In future work, it also will be beneficial to incorporate attention mechanism, which is gaining momentum in other domain adaptation tasks, in producing higher quality images and reduce impact of speckles.

{\small
\bibliographystyle{ieee_fullname}
\bibliography{PaperForReview}

\begin{thebibliography}{1}\itemsep=-1pt

\bibitem{anderson2017earth}
Katherine Anderson, Barbara Ryan, William Sonntag, Argyro Kavvada, and Lawrence
  Friedl.
\newblock Earth observation in service of the 2030 agenda for sustainable
  development.
\newblock {\em Geo-spatial Information Science}, 20(2):77--96, 2017.

\bibitem{bermudez2019synthesis}
Jose~D Bermudez, Patrick~N Happ, Raul~Q Feitosa, and Dario~AB Oliveira.
\newblock Synthesis of multispectral optical images from sar/optical
  multitemporal data using conditional generative adversarial networks.
\newblock {\em IEEE Geoscience and Remote Sensing Letters}, 16(8):1220--1224,
  2019.

\bibitem{cha2022multiearth}
Miriam Cha, Kuan~Wei Huang, Morgan Schmidt, Gregory Angelides, Mark Hamilton,
  Sam Goldberg, Armando Cabrera, Phillip Isola, Taylor Perron, Bill Freeman,
  et~al.
\newblock Multiearth 2022--multimodal learning for earth and environment
  workshop and challenge.
\newblock {\em arXiv preprint arXiv:2204.07649}, 2022.

\bibitem{chuvieco2010advances}
Emilio Chuvieco, Jonathan Li, and Xiaojun Yang.
\newblock {\em Advances in earth observation of global change}.
\newblock Springer, 2010.

\bibitem{fuentes2019sar}
Mario Fuentes~Reyes, Stefan Auer, Nina Merkle, Corentin Henry, and Michael
  Schmitt.
\newblock Sar-to-optical image translation based on conditional generative
  adversarial networks—optimization, opportunities and limits.
\newblock {\em Remote Sensing}, 11(17):2067, 2019.

\bibitem{isola2017image}
Phillip Isola, Jun-Yan Zhu, Tinghui Zhou, and Alexei~A Efros.
\newblock Image-to-image translation with conditional adversarial networks.
\newblock In {\em Proceedings of the IEEE conference on computer vision and
  pattern recognition}, pages 1125--1134, 2017.

\bibitem{turnes2020atrous}
Javier~Noa Turnes, Jose David~Bermudez Castro, Daliana~Lobo Torres, Pedro
  Juan~Soto Vega, Raul~Queiroz Feitosa, and Patrick~N Happ.
\newblock Atrous cgan for sar to optical image translation.
\newblock {\em IEEE Geoscience and Remote Sensing Letters}, 19:1--5, 2020.

\bibitem{zhu2017unpaired}
Jun-Yan Zhu, Taesung Park, Phillip Isola, and Alexei~A Efros.
\newblock Unpaired image-to-image translation using cycle-consistent
  adversarial networks.
\newblock In {\em Proceedings of the IEEE international conference on computer
  vision}, pages 2223--2232, 2017.

\end{thebibliography}
}

\end{document}